\documentclass[runningheads]{llncs}

\usepackage{graphicx}
\usepackage{subfigure}

\usepackage{hyperref}

\usepackage[capitalise,nameinlink]{cleveref}
\usepackage{iflang}
\IfLanguageName{ngerman}{
  \crefname{table}{Tab.}{Tab.}
  \Crefname{table}{Tabelle}{Tabellen}
  \crefname{figure}{\figurename}{\figurename}
  \Crefname{figure}{Abbildung}{Abbildungen}
  \crefname{equation}{Gleichung}{Gleichungen}
  \Crefname{equation}{Gleichung}{Gleichungen}
  \crefname{listing}{\lstlistingname}{\lstlistingname}
  \Crefname{listing}{Listing}{Listings}
  \crefname{section}{Abschnitt}{Abschnitte}
  \Crefname{section}{Abschnitt}{Abschnitte}
  \crefname{paragraph}{Abschnitt}{Abschnitte}
  \Crefname{paragraph}{Abschnitt}{Abschnitte}
  \crefname{subparagraph}{Abschnitt}{Abschnitte}
  \Crefname{subparagraph}{Abschnitt}{Abschnitte}
}{
  \crefname{section}{Sect.}{Sect.}
  \Crefname{section}{Section}{Sections}
  \crefname{listing}{\lstlistingname}{\lstlistingname}
  \Crefname{listing}{Listing}{Listings}
}

\usepackage{booktabs}
\usepackage{multirow}

\usepackage{bbding}

\usepackage{amsfonts}

\usepackage{pifont}
\newcommand{\cmark}{\ding{51}}%
\newcommand{\xmark}{\ding{55}}%

\begin{document}
\title{LightCAKE: A Lightweight Framework for Context-Aware Knowledge Graph Embedding}

\titlerunning{LightCAKE}

\author{Zhiyuan Ning\inst{1,2} \and
Ziyue Qiao\inst{1,2} \and
Hao Dong\inst{1,2} \and
Yi Du\inst{1(}\Envelope\inst{)} \and
Yuanchun Zhou\inst{1}}

\authorrunning{Z. Ning et al.}

\institute{Computer Network Information Center,\\
 Chinese Academy of Sciences, Beijing, China \and
 University of Chinese Academy of Sciences, Beijing, China
\email{\{ningzhiyuan,qiaoziyue,donghao,duyi,zyc\}@cnic.cn}}

\maketitle              

\begin{abstract}
Knowledge graph embedding (KGE) models learn to 
project symbolic entities and relations into a continuous vector space based on the observed triplets. 
However, existing KGE models cannot make a proper trade-off between the graph context and the model complexity, 
which makes them still far from satisfactory. 
In this paper, we propose a lightweight framework named LightCAKE for context-aware KGE. 
LightCAKE explicitly models the graph context without introducing redundant trainable parameters, 
and uses an iterative aggregation strategy to integrate the context information into the entity/relation embeddings. 
As a generic framework, it can be used with many simple KGE models to achieve excellent results. 
Finally, extensive experiments on public benchmarks demonstrate the efficiency and effectiveness of our framework.

\keywords{Knowledge graph embedding  \and Lightweight \and Graph context}
\end{abstract}
\section{Introduction}\label{sec:introduction}
Recently, large-scale knowledge graphs (KGs) have been widely applied to numerous AI-related applications. 
Indeed, KGs are usually expressed as multi-relational directed graphs composed of entities as nodes and relations as edges. 
The real-world facts stored in KGs are modeled as triplets (head entity, relation, tail entity), which are denoted as $(h, r, t)$.

Nevertheless, KGs are usually incomplete due to the constant emergence of new knowledge. 
To address this issue, a series of knowledge graph embedding (KGE) models have been proposed~\cite{wang2017knowledge}. 
KGE models project symbolic entities and relations into a continuous vector space, 
and use scoring functions to measure the plausibility of triplets. 
By optimizing the scoring functions to assign higher scores to true triplets than invalid ones, 
KGE models learn low-dimensional representations (called embeddings) for all entities and relations, 
and these embeddings are then used to predict new facts. 
Most of the previous KGE models use translation distance based~\cite{bordes2013translating,wang2014knowledge} 
and semantic matching based~\cite{trouillon2016complex,yang2015embedding} scoring functions 
which perform additive and multiplicative operations, respectively. 
These models have been shown to be scalable and effective.

\begin{figure}[htbp]
  \centering
  \subfigure[Entity Context]{
  \label{fig:ec}
  \centering
  \includegraphics[width=0.478\textwidth]{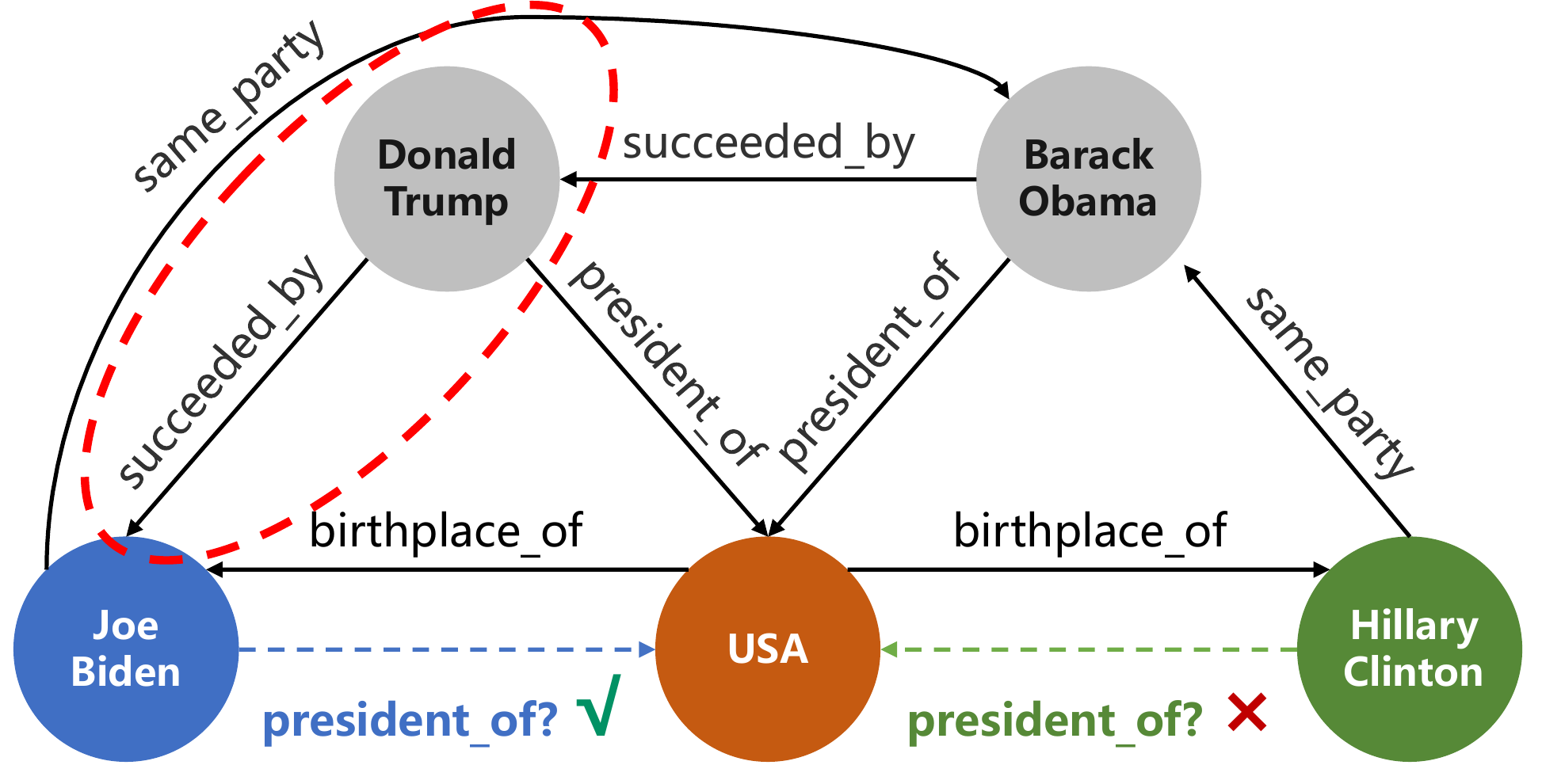}}
  \subfigure[Relation Context]{
  \label{fig:rc}
  \centering
  \includegraphics[width=0.478\textwidth]{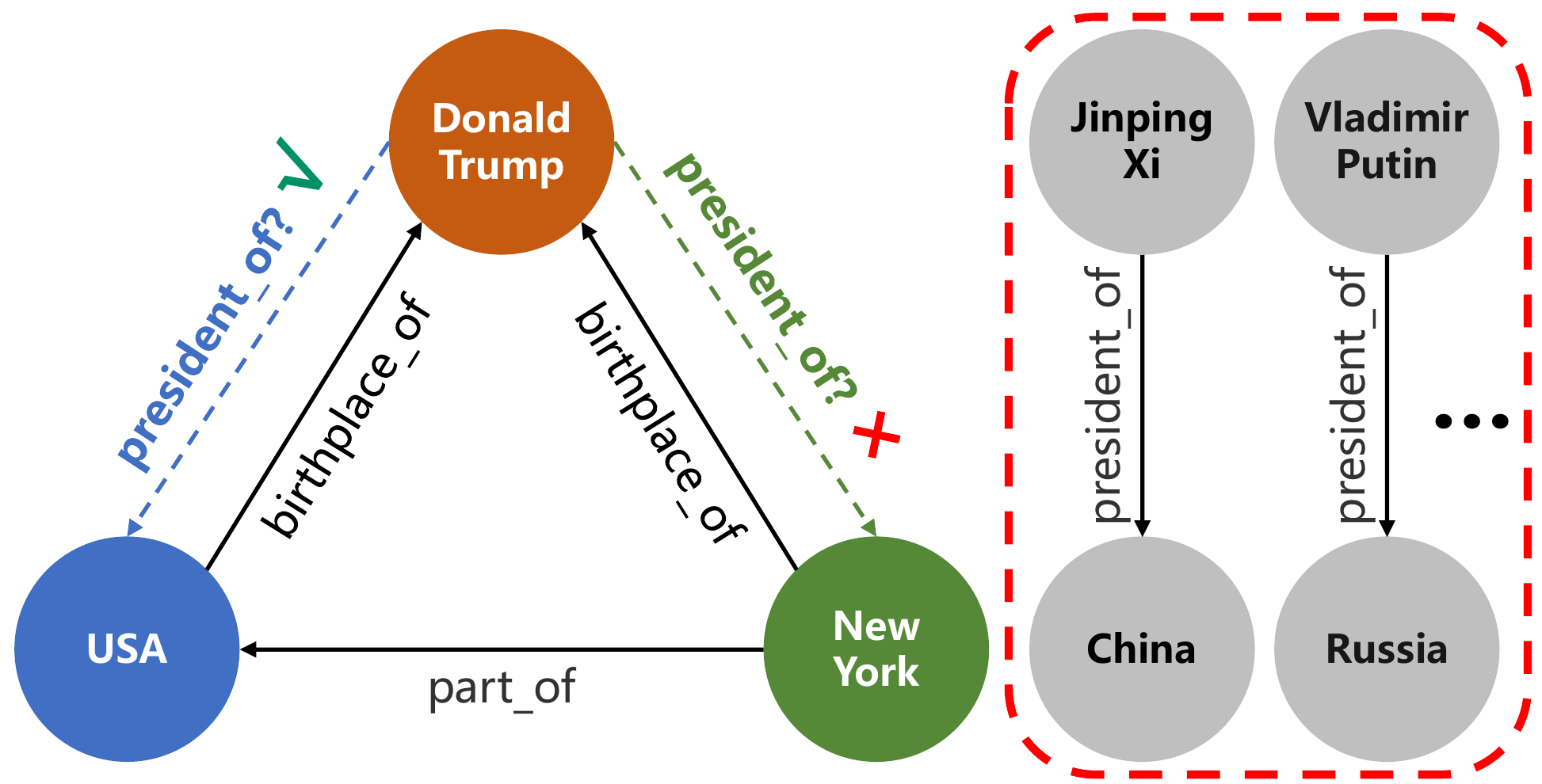}}
  \caption{\textbf{Examples of graph context which can help the relation prediction in knowledge graph.}
Nodes represent entities, solid lines represent actual relations, dashed lines represent the relations to be predicted.
Red dashed boxes frame the critical entity context (Figure a) and relation context (Figure b) that
can provide important information for correctly predicting the relation between two entities.}
  \label{fig:context}
\end{figure}

However, the aforementioned KGE models only focus on modeling individual triplets and ignore the \textbf{graph context}, 
which contains plenty of valuable structural information. 
We argue that there are two types of important graph contexts required for successfully predicting the relation between two entities: 
(1) The \textbf{entity context}, i.e., for an entity, 
its neighboring nodes and the corresponding edges connecting the entity to its neighboring nodes. 
The entity context depicts the subtle differences between two entities. 
As an example shown in \cref{fig:ec}, we aim to predict whether \textit{Joe Biden} or \textit{Hillary Clinton} 
is the president of the \textit{USA}. 
Both of them have the same relation "\textit{birthplace\rule[-2pt]{0.15cm}{0.3pt}of}" with the \textit{USA}, 
but they have distinct entity contexts. 
\textit{Joe Biden's} neighboring node, \textit{Donald Trump}, is the president of the \textit{USA}, 
and \textit{Biden} is his successor. 
Whereas there is no such relationship between \textit{Hillary Clinton} and her neighboring nodes. 
Capturing such entity context will help predict the correct triplet 
(\textit{Joe Biden}, \textit{president\rule[-2pt]{0.15cm}{0.3pt}of}, \textit{USA}). 
(2) The \textbf{relation context}, i.e., the two endpoints of a given relation. 
Relation context implicitly indicates the category of related entities. 
Taking \cref{fig:rc} as an example, 
both the \textit{USA} and \textit{New York} were \textit{Donald Trump's} birthplace, 
but according to the  context of "\textit{president\rule[-2pt]{0.15cm}{0.3pt}of}", 
the related tail entities \{\textit{China}, \textit{Russia}, \dots\} tend to be a set of countries. 
Since \textit{New York} is a city and it is part of the \textit{USA} which is a country, 
(\textit{Donald Trump}, \textit{president\rule[-2pt]{0.15cm}{0.3pt}of}, \textit{USA}) is the right triplet. 
Moreover, entities and relations rarely appear in isolation, 
so considering entity context and relation context together will provide more beneficial information.

In order to model the graph context, 
some recent work has attempted to apply graph neural network (GNN) to KGE~\cite{bansal2019a2n,schlichtkrull2018modeling}. 
These GNN-based KGE models are effective to aggregate information from multi-hop neighbors to enrich the entity/relation representation. 
However, GNN introduces more model parameters and tensor computations, 
therefore making it difficult to utilize these models for large-scale real-world KGs. 
In addition, most GNN-based KGE models only exploit entity context or relation context individually, 
which may lead to information loss.

In this paper, we propose a \textbf{Light}weight Framework for 
\textbf{C}ontext-\textbf{A}ware \textbf{K}nowledge Graph \textbf{E}mbedding 
(\textbf{LightCAKE}) to address the shortcomings of existing models. 
LightCAKE first builds the context star graph to model the entity/relation context. 
It then uses non-parameterized operations like 
subtraction (inspired by TransE\cite{bordes2013translating}) or multiplication (inspired by DistMult\cite{yang2015embedding}) 
to encode context nodes in the context star graph. 
Lastly, every entity/relation node in the context star graph aggregates information from its surrounding context nodes 
based on the weights calculated by a scoring function. 
LightCAKE considers both entity context and relation context, 
and introduces no new parameters, making it very lightweight and capable of being used on large-scale KGs. 
The contributions of our work can be summarized as follows: 
(1) We propose a lightweight framework (LightCAKE) for KGE that 
explicitly model the entity context and relation context without the sacrifice in the model complexity; 
(2) As a general framework, we can apply many simple methods 
like TransE\cite{bordes2013translating} and DistMult\cite{yang2015embedding} to LightCAKE; 
(3) Through extensive experiments on relation prediction task, we demonstrate the effectiveness and efficiency of LightCAKE.

\section{Related Work}\label{sec:related_work}
Most early KGE models only exploit the triplets and 
can be roughly categorized into two classes\cite{wang2017knowledge}: 
translation distance based and semantic matching based. 
Translation distance based models are also known as \textbf{additive models}, 
since they project head and tail entities into the same embedding space, 
and treat the relations as the translations from head entities to tail entities. 
The objective is that the translated head entity should be close to the tail entity. 
TransE~\cite{bordes2013translating} is the first and most representative of such models. 
A series of work is conducted along this line such as TransR~\cite{lin2015learning} and TransH~\cite{wang2014knowledge}. 
On the other hand, 
semantic matching based models such as DistMult~\cite{yang2015embedding} and ComplEx~\cite{trouillon2016complex} 
use multiplicative score functions for computing the plausibility of the given triplets, 
so they are also called \textbf{multiplicative models}. 
Both models are conceptually simple and it is easy to apply them to large-scale KGs. 
But they ignore the structured information stored in the graph context of KGs.

In contrast, \textbf{GNN-based} models attempt to use GNN for graph context modeling. 
These models first aggregate graph context into entity/relation embeddings through GNN, 
then pass the context-aware embeddings to the context-independent scoring functions for scoring. 
R-GCN~\cite{schlichtkrull2018modeling} 
is an extension of the graph convolutional network~\cite{kipf2017semi} on relational data. 
It applies a convolution operation to the neighboring nodes of each entity and assigns them equal weights. 
A2N~\cite{bansal2019a2n} uses a method similar to graph attention networks~\cite{velickovic2018graph} 
to further distinguish the weights of neighboring nodes.
However, this type of KGE models suffer from overparameterization 
since there are many parameters in GNN, 
which will hinder the application of such models to large-scale KGs. 
In addition, they don't integrate entity context and relation context, which may cause information loss.

\begin{figure}[htbp]
  \centering
  \includegraphics[width=1\textwidth]{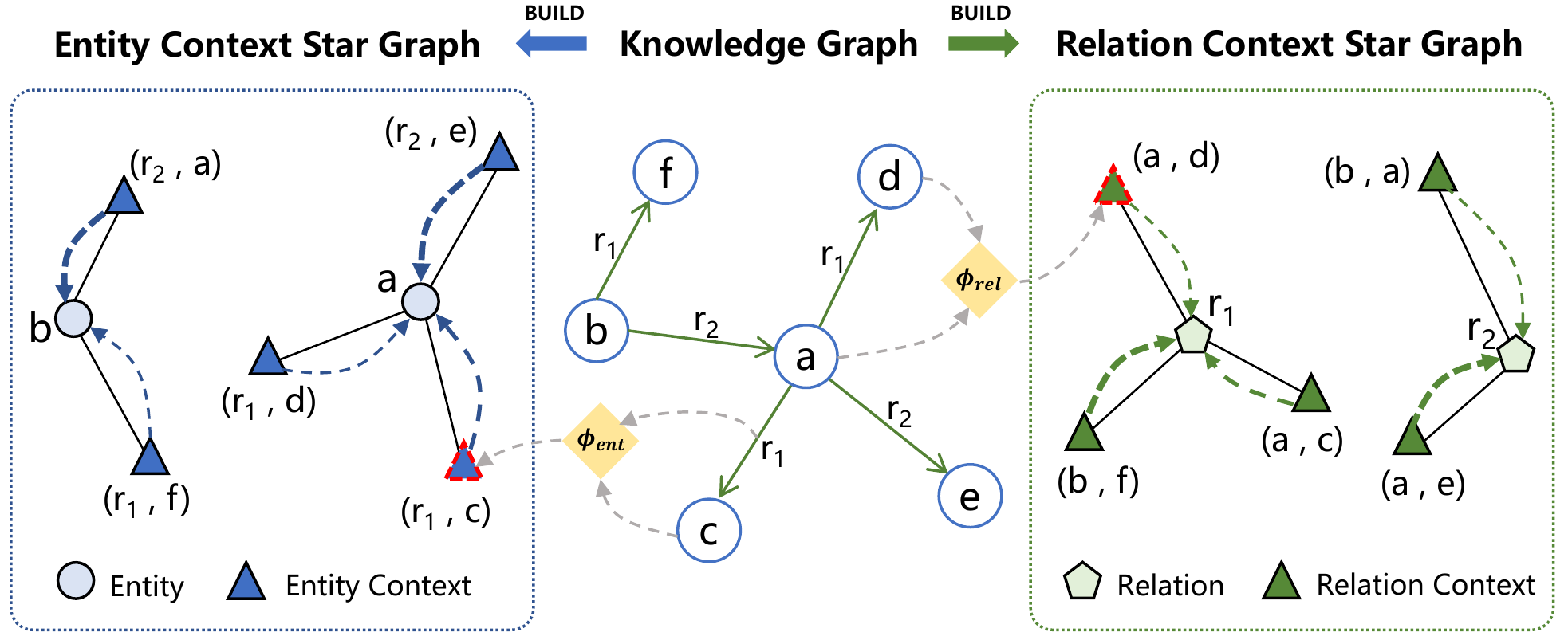}
  \caption{\textbf{Overview of LightCAKE.}
  (1) For a KG (Middle), we build an entity context star graph (Left) for all entities
  and a relation context star graph (Right) for all relations.
  In entity/relation context star graph,
  each entity/relation is surrounded by its entity/relation context and they are connected to each other by solid black lines.
  (2) The yellow rhombus $\phi_{ent}$ and $\phi_{rel}$ denote context encoders
  (Details in \cref{sec:lightcake_details}),
  and the gray dashed line indicates the input and output of the encoders.
  (3) The blue dashed line denotes the weight $\alpha$ (\cref{con:alpha}), 
  and the green dashed line denotes the weight $\beta$ (\cref{con:beta}). 
  The thicker the line, the greater the weight.
  }
  \label{fig:fw}
\end{figure}

\section{Preliminaries}\label{sec:preliminaries}
\subsection{Notation and Problem Formulation}\label{sec:notation_and_problem_formulation}
A KG can be considered as a collection of triplets 
$\mathcal { G } = \{ ( h , r , t ) \mid ( h , r , t ) \in \mathcal { E } \times \mathcal { R } \times \mathcal { E } \}$, 
where $\mathcal { E }$ is the entity set and $\mathcal { R }$ is the relation set. 
$h, t \in \mathcal { E }$ represent the head entity and tail entity, 
$r \in \mathcal { R }$ denotes the relation linking from the head entity $h$ to tail entity $t$. 
Given a triplet $(h, r, t)$, the corresponding embeddings are $e_{h}, e_{r}, e_{t}$, 
where $e_{h}, e_{r}, e_{t} \in \mathbb{R}^{d}$, and $d$ is the embedding dimension. 
KGE models usually define a scoring function 
$\psi: \mathbb{R}^{d} \times \mathbb{R}^{d} \times \mathbb{R}^{d} \to \mathbb{R}$. 
It takes the corresponding embedding $(e_{h}, e_{r}, e_{t})$ of a triplet $(h, r, t)$ as input, 
and produces a score reflecting the plausibility of the triplet. 

In this paper, the objective is to predict the missing links in $\mathcal { G }$, i.e., 
given an entity pair $(h, t)$, we aim to predict the missing relation $r$ between them. 
We refer to this task as \textbf{relation prediction}. 
Some related work formulates this problem as link prediction, i.e., 
predicting the missing tail/head entity given a head/tail entity and a relation. 
The two problems have proven to be actually reducible to each other~\cite{wang2020entity}.

\subsection{Context Star Graph}\label{sec:context_star_graph}
\begin{definition}
\textbf{Entity Context}: 
For an entity $h$ in $\mathcal { G }$, the entity context of $h$ is defined as 
$\mathcal {C}_{ent}(h) = \{(r, t) \mid (h, r, t) \in \mathcal { G }\}$, i.e., 
all the (relation, tail) pairs in $\mathcal { G }$ whose head is $h$. 
\end{definition}

\begin{definition}
\textbf{Relation Context}: 
For a relation $r$ in $\mathcal { G }$, the relation context of $r$ is defined as 
$\mathcal {C}_{rel}(r) = \{(h, t) \mid (h, r, t) \in \mathcal { G }\}$, i.e., 
all the (head, tail) pairs in $\mathcal { G }$ whose relation is $r$. 
\end{definition}

Note that the entity context $\mathcal {C}_{ent}(h)$ only considers 
the neighbors of $h$ for its outgoing edges and ignores the neighbors for its incoming edges. 
This is because for each triplet $(h, r, t) \in \mathcal { G }$, 
we create a corresponding inverse triplet $(t, r^{-1}, h)$ and add it to $\mathcal { G }$. 
In this way, for entity $t$,  $\{(r, h) \mid (h, r, t) \in \mathcal { G }\}$ can be converted to a format of 
$\{(r^{-1}, h) \mid (t, r^{-1}, h) \in \mathcal { G }\}$, and it is equivalent to  $\mathcal {C}_{ent}(t)$. 
Thus, $\mathcal {C}_{ent}(\cdot)$ can contain both the outgoing and incoming neighbors for each entity.

To explicitly model entity context and relation context 
for a KG $\mathcal { G }$ (As shown in \cref{fig:fw} middle), 
we construct an \textbf{entity context star graph} (As shown in \cref{fig:fw} left) 
and a \textbf{relation context star graph} (As shown in \cref{fig:fw} right), respectively. 
In the entity context star graph, all the central nodes are the entities in $\mathcal { G }$, 
and each entity $h$ is surrounded by its entity context $\mathcal {C}_{ent}(h)$. 
Similarly, 
in the relation context star graph, all the central nodes are the relations in $\mathcal { G }$, 
and each relation $r$ is surrounded by its relation context $\mathcal {C}_{rel}(r)$.

\section{Methodology}\label{sec:methodology}
Given the context star graph, 
LightCAKE can (1) encode each entity/relation context node into an embedding; 
(2) learn the context-aware embedding for each entity/relation 
by iteratively aggregating information from its context nodes. 

\subsection{LightCAKE Details}\label{sec:lightcake_details}
Denote $e_{h}^{(0)}$ and $e_{r}^{(0)}$ as the randomly initialized embedding of 
an entity $h$ and a relation $r$ respectively. The aggregation functions are formulated as:
\begin{equation}
  e_{h}^{(l+1)}=e_{h}^{(l)}+\sum_{(r',t')\in\mathcal{C}_{ent}(h)}\alpha_{h,(r',t')}^{(l)}\phi_{ent}(e_{r'},e_{t'})
  \label{con:general_eh}
\end{equation}
\begin{equation}
  e_{r}^{(l+1)}=e_{r}^{(l)}+\sum_{(h',t')\in\mathcal{C}_{rel}(r)}\beta_{r,(h',t')}^{(l)}\phi_{rel}(e_{h'},e_{t'})
  \label{con:general_er}
\end{equation}
Here, $e_{h}^{(l+1)}$ and $e_{r}^{(l+1)}$ are the embeddings of $h$ and $r$ after $l$-iterations aggregations. 
$0 \leq l \leq L$ and $L$ is the total number of iterations. 
$\phi_{ent}(\cdot): \mathbb{R}^{d} \times \mathbb{R}^{d} \to \mathbb{R}^{d}$ is the entity context encoder,  
and $\phi_{rel}(\cdot): \mathbb{R}^{d} \times \mathbb{R}^{d} \to \mathbb{R}^{d}$ is the relation context encoder. 
$\alpha_{h,(r',t')}^{(l)}$ and $\beta_{r,(h',t')}^{(l)}$ 
are the weights in iteration $l$, 
representing how important each context node is for $h$ and $r$, respectively. 
We introduce the scoring function $\psi(\cdot)$ to calculate them: 
\begin{equation}
  \alpha_{h,(r',t')}^{(l)}=\frac{\mathrm{exp}(\psi(e_{h}^{(l)},e_{r^{'}}^{(l)},e_{t^{'}}^{(l)})}{\sum_{(r^{''},t^{''})\in\mathcal{C}_{ent}(h)}\mathrm{exp}(\psi(e_{h}^{(l)},e_{r^{''}}^{(l)},e_{t^{''}}^{(l)}))}
  \label{con:alpha}
\end{equation}
\begin{equation}
  \beta_{r,(h',t')}^{(l)}=\frac{\mathrm{exp}(\psi(e_{h^{'}}^{(l)},e_{r}^{(l)},e_{t^{'}}^{(l)})}{\sum_{(h^{''},t^{''})\in\mathcal{C}_{rel}(r)}\mathrm{exp}(\psi(e_{h^{''}}^{(l)},e_{r}^{(l)},e_{t^{''}}^{(l)}))}
  \label{con:beta}
\end{equation}
When \cref{con:general_eh} and \cref{con:general_er} are iteratively executed $L$ times, 
for any $h, t \in \mathcal { E }$ and $r \in \mathcal { R }$, 
we obtain the final context-enhanced embeddings $e_{h}^{(L)},e_{r}^{(L)},e_{t}^{(L)}$. 
To perform relation prediction, we compute the probability of the relation $r$ 
given the head entity $h$ and tail entity $t$ using a $\mathrm{softmax}$ function:
\begin{equation}
  p(r|h,t) =\frac{\mathrm{exp}(\psi(e_{h}^{(L)},e_{r}^{(L)},e_{t}^{(L)})}{\sum_{r^{'}\in\mathcal{R}}\mathrm{exp}(\psi(e_{h}^{(L)},e_{r^{'}}^{(L)},e_{t}^{(L)}))}
  \label{con:final_p}
\end{equation}
where $\mathcal{R}$ is the set of relations, 
$\psi(\cdot)$ is the same scoring function used in \cref{con:alpha} and \cref{con:beta}. 
Then, we train the model by minimizing the following loss function: 
\begin{equation}
  \mathcal{L}=-\frac{1}{|\mathcal{D}|}\sum_{i=0}^{|\mathcal{D}|}\log p\left(r_{i}\mid h_{i},t_{i}\right)
\end{equation}
where $\mathcal{D}$ is the training set, and $(h_{i},r_{i},t_{i})\in\mathcal{D}$ is one of the training triplets.

\subsection{Special Cases of LightCAKE}\label{special_cases_of_lightCAKE}
LightCAKE is a generic framework, 
and we can substitute different scoring function $\psi(\cdot)$ 
of different KGE models into \cref{con:alpha}, \cref{con:beta}, and \cref{con:final_p}. 
And we can design different $\phi_{ent}(\cdot)$ and $\phi_{rel}(\cdot)$ to encode context. 
In order to make the framework lightweight, 
we apply TransE~\cite{bordes2013translating} and DistMult~\cite{yang2015embedding}, 
which are the simplest and most representative of the additive models and multiplicative models respectively, 
to LightCAKE. 

\subsubsection{LightCAKE-TransE}\label{sec:lightcake_transe}
The scoring function of TransE~\cite{bordes2013translating} is:
\begin{equation}
  \psi_{TransE}(e_{h}, e_{r}, e_{t}) = - \left\| e_{h} + e_{r} - e_{t} \right\|_{2} = - \left\| e_{t} - e_{r} - e_{h} \right\|_{2}
  \label{con:sf_transe}
\end{equation}
where $\left\| \cdot \right\|_{2}$ is the L2-norm. \cref{con:sf_transe} can be decomposed of the two following steps:
\begin{equation}
  e_{(h,r,t)} = \mathcal{V}_{TransE}(e_{h}, e_{r}, e_{t}) = e_{t} - e_{r} - e_{h}
  \label{con:v_transe}
\end{equation}
\begin{equation}
  score = \mathcal{S}_{TransE}(e_{(h,r,t)}) = - \left\| e_{(h,r,t)} \right\|_{2}
\end{equation}
where $\mathcal{V_{\cdot}}: \mathbb{R}^{d} \times \mathbb{R}^{d} \times \mathbb{R}^{d} \to \mathbb{R}^{d}$ and 
$\mathcal{S_{\cdot}}: \mathbb{R}^{d} \to \mathbb{R}$. 
The $e_{(h,r,t)}$ denotes the embedding of a triplet $(h,r,t)$,
$score$ denotes the score of the triplet.
In \cref{con:v_transe}, TransE uses addition and subtraction to encode triplets. 
Moreover, the operation between $e_{r}$ and $e_{t}$ is subtraction, 
and the operation between $e_{h}$ and $e_{t}$ is also subtraction.
So we design $\phi_{ent}(e_{r'}, e_{t'}) = e_{t'} - e_{r'}$ 
and $\phi_{rel}(e_{h'}, e_{t'}) = e_{t'} - e_{h'}$ to encode context, 
then the aggregation function of LightCAKE-TransE can be formalized as:
\begin{equation}
  e_{h}^{(l+1)}=e_{h}^{(l)}+\sum_{(r',t')\in\mathcal{C}_{ent}(h)}\alpha_{h,(r',t')}^{(l)}(e_{t'} - e_{r'})
\end{equation}
\begin{equation}
  e_{r}^{(l+1)}=e_{r}^{(l)}+\sum_{(h',t')\in\mathcal{C}_{rel}(r)}\beta_{r,(h',t')}^{(l)}(e_{t'} - e_{h'})
\end{equation}
Lastly, substitute $\psi_{TransE}(e_{h}, e_{r}, e_{t})$ from \cref{con:sf_transe} into \cref{con:alpha}, \cref{con:beta} and \cref{con:final_p}, 
we will get the complete LightCAKE-TransE.

\subsubsection{LightCAKE-DistMult}\label{sec:lightcake-distmult} 
The scoring function of DistMult~\cite{yang2015embedding} is:
\begin{equation}
  \psi_{DistMult}(e_{h}, e_{r}, e_{t}) = \langle e_{h} , e_{r}, e_{t} \rangle
  \label{con:sf_distmult}
\end{equation}
where $\langle \cdot \rangle$ denotes the generalized dot product.
\cref{con:sf_distmult} can be decomposed of the two following steps:
\begin{equation}
  e_{(h,r,t)} = \mathcal{V}_{DistMult}(e_{h}, e_{r}, e_{t}) = e_{h} \odot e_{r} \odot e_{t}
  \label{con:v_distmult}
\end{equation}
\begin{equation}
  score = \mathcal{S}_{DistMult}(e_{(h,r,t)}) = \sum_{i} e_{(h,r,t)}[i]
\end{equation}
where $\odot$ denotes the element-wise product, 
and $e_{(h,r,t)}[i]$ denotes the i-th element in embedding $e_{(h,r,t)}$.
In \cref{con:v_distmult}, DistMult uses multiplication to encode triplets.
Moreover, the operation between $e_{r}$ and $e_{t}$ is multiplication, 
and the operation between $e_{h}$ and $e_{t}$ is also multiplication.
So we design $\phi_{ent}(e_{r'}, e_{t'}) = e_{t'} \odot e_{r'}$ 
and $\phi_{rel}(e_{h'}, e_{t'}) = e_{t'} \odot e_{h'}$ to encode context, 
then the aggregation function of LightCAKE-DistMult can be formalized as:
\begin{equation}
  e_{h}^{(l+1)}=e_{h}^{(l)}+\sum_{(r',t')\in\mathcal{C}_{ent}(h)}\alpha_{h,(r',t')}^{(l)}(e_{t'} \odot e_{r'})
\end{equation}
\begin{equation}
  e_{r}^{(l+1)}=e_{r}^{(l)}+\sum_{(h',t')\in\mathcal{C}_{rel}(r)}\beta_{r,(h',t')}^{(l)}(e_{t'} \odot e_{h'})
\end{equation}
Lastly, substitute $\psi_{DistMult}(e_{h}, e_{r}, e_{t})$ from \cref{con:sf_distmult} into \cref{con:alpha}, \cref{con:beta} and \cref{con:final_p}, 
we will get the complete LightCAKE-DistMult.

Notably, there are no extra trainable parameters introduced in 
LightCAKE-TransE and LightCAKE-DistMult, 
making them lightweight and efficient.

\section{Experiments}\label{sec:experiments}

\subsection{Dataset}\label{sec:dataset}
We evaluate LightCAKE on four popular benchmark datasets WN18RR~\cite{dettmers2018conve}, 
FB15K-237~\cite{toutanova2015observed}, NELL995~\cite{xiong2017deeppath} and DDB14~\cite{wang2020entity}. 
WN18RR is extracted from WordNet, containing conceptual-semantic and lexical relations among English words. 
FB15K-237 is extracted from Freebase, a large-scale KG with general human knowledge. 
NELL995 is extracted from the 995th iteration of the NELL system containing general knowledge. 
DDB14 is extracted from the Disease Database, 
a medical database containing terminologies and concepts as well as their relationships. 
The statistics of the datasets are summarized in \cref{statistics}.

\begin{table}[htbp]
  \centering
  \caption{\textbf{Statistics of four datasets.}
  avg.$|\mathcal {C}_{ent}(h)|$ and avg.$|\mathcal {C}_{rel}(r)|$ 
  represent the average number of entity context and relation context, respectively.}
  \scalebox{0.9}[0.9]{
  \begin{tabular}{lcccc}
  \toprule
  Dataset  & FB15K-237          & WN18RR               & NELL995              & DDB14                \\ \midrule
  \#entitiy  & 14,541              & 40,943               & 63,917               & 9,203                \\
  \#relation      & 237          & 11                   & 198                  & 14                   \\ \midrule
  \#train      & 272,115           & 86,835               & 137,465              & 36,561               \\
  \#test     & 17,535          & 3,034                & 5,000                & 4,000                \\
  \#valid    & 20,466             & 3,134                & 5,000                & 4,000                \\ \midrule
  avg.$|\mathcal {C}_{ent}(h)|$  & 37.4       & 4.2       & 4.3      & 7.9     \\
  avg.$|\mathcal {C}_{rel}(r)|$   & 1148.2     & 7894.1       & 694.3        & 2611.5	  \\ \bottomrule
  \end{tabular}
  }
  \label{statistics}
\end{table}

\subsection{Baselines}\label{sec:Baselines}
To prove the effectiveness of LightCAKE, 
we compare LightCAKE-TransE and LightCAKE-DistMult with six baselines, 
including (1) original TransE and DistMult without aggregating entity context and relation context; 
(2) three state-of-the-art KGE models: ComplEx, SimplE, RotatE;  
(3) a classic GNN-based KGE model: R-GCN. 
Brief descriptions of baselines are as follows:

\textbf{TransE~\cite{bordes2013translating}:} TransE is one of the most widely-used KGE models 
which translates the head embedding into tail embedding by adding it to relation embedding. 

\textbf{DistMult~\cite{yang2015embedding}:} DistMult is a popular tensor factorization based model 
which uses a bilinear score function to compute scores of knowledge triplets. 

\textbf{ComplEx~\cite{trouillon2016complex}:} ComplEx is an extension of DistMult which 
embeds entities and relations into complex vectors instead of real-valued ones. 

\textbf{SimplE~\cite{kazemi2018simple}:} SimplE is a simple interpretable fully-expressive 
tensor factorization model for knowledge graph completion.

\textbf{RotatE~\cite{sun2018rotate}:} RotatE defines each relation as a rotation from 
the head entity to the tail entity in the complex vector space.

\textbf{R-GCN~\cite{schlichtkrull2018modeling}:} RGCN is a variation of graph neural network, 
it can deal with the highly multi-relational knowledge graph data and 
aggregate context information to entities. 

To simplify, we use $\mathcal{L}$-TransE to represent LightCAKE-TransE 
and use $\mathcal{L}$-DistMult to represent LightCAKE-DistMult.

\subsection{Experimental Settings}\label{sec:experimental_settings}
We use Adam~\cite{kingma2015adam} as the optimizer with the learning rate as 5e-3. 
We set the embedding dimension of entity and relation as 256, $l_{2}$ penalty coefficient as 1e-7, batch size as 512, 
the total number of iterations $L$ as 4 and a maximum of 20 epochs. 
Moreover, we use early stopping for training, 
and all the training parameters are randomly initialized.

We evaluate all methods in the setting of relation prediction, 
i.e., for a given entity pair $(h, t)$ in the test set, 
we rank the ground-truth relation type $r$ against all other candidate relation types. 
We compare our models with baselines using the following metrics: 
(1) Mean Reciprocal Rank (MRR, the mean of all the reciprocals of predicted ranks); 
(2) Mean Rank (MR, the mean of all the predicted ranks); 
(3) Hit@3(the proportion of correctly predicted entities ranked in the top 3 predictions). 

\begin{table}[htbp]
  \caption{\textbf{Results of relation prediction. (Bold: best; Underline: runner-up.)} 
  The results of ComplEx, SimplE and RotatE are taken from \cite{wang2020entity}.
  Noted that the trainable parameters in $\mathcal{L}$-TransE and $\mathcal{L}$-DistMult 
  are only entity embeddings and relation embeddings, for a fair comparison, 
  we only choose those 3 traditional baselines from \cite{wang2020entity} with a small number of parameters. 
  In addition, in order to compare context-aware KGE and context-independent KGE in the same experimental environment 
  to prove the validity of LightCAKE, we implemented TransE and DistMult ourselves.}
  \centering
  \scalebox{0.85}{
  \begin{tabular}{l|ccc|ccc|ccc|ccc}
  \toprule
      \multirow{3}{*}{\large{Method}}       & \multicolumn{3}{c|}{WN18RR}                                                   & \multicolumn{3}{c|}{FB15K-237} 
                                              & \multicolumn{3}{c|}{NELL995}               & \multicolumn{3}{c}{DDB14}       \\ \cmidrule{2-13}
           & \multicolumn{1}{c}{MRR} & \multicolumn{1}{c}{MR$\downarrow$} & \multicolumn{1}{c|}{Hit@3} & \multicolumn{1}{c}{MRR} & \multicolumn{1}{c}{MR$\downarrow$} & \multicolumn{1}{c|}{Hit@3}
           & \multicolumn{1}{c}{MRR} & \multicolumn{1}{c}{MR$\downarrow$} & \multicolumn{1}{c|}{Hit@3} & \multicolumn{1}{c}{MRR} & \multicolumn{1}{c}{MR$\downarrow$} & \multicolumn{1}{c}{Hit@3} \\ \midrule
      
  ComplEx  & 0.840                   & 2.053                  & 0.880                      & 0.924                   & 1.494                  & 0.970   
      & 0.703                   & 23.040                  &  0.765                      & 0.953                   & 1.287                  &  0.968  \\
  SimplE   & 0.730                   & 3.259                  & 0.755                      & \textbf{0.971}         & 1.407                  & \underline{0.987}  
      & 0.716                  &  26.120                  & 0.748                      & 0.924               & 1.540                  & 0.948 \\
  RotatE   & 0.799                   & 2.284                  & 0.823                      & \underline{0.970}                  & \underline{1.315}             & 0.980      
       & 0.729                  &  23.894                  &  0.756                     & 0.953                 & 1.281                  &  0.964 \\   \midrule
  RGCN   & 0.823                   & 2.144                  & 0.854                      &  0.954                  &  1.498                 &0.973       &0.731                    &22.917                   &0.749                       & 0.951                   &  1.278                 & 0.965               \\ \midrule
  
  TransE   &0.789                    &1.755                   &0.918                       & 0.932                   &1.979                  &0.952           &0.719                    &16.654                   &0.766                       &0.936                    &  1.487                 &  0.957          \\
  $\mathcal{L}$-TransE   &0.813                    &\underline{1.648}       &\underline{0.933}                       &0.943                    &2.281                    &0.962          &\underline{0.793}        &\underline{9.325}                   &\underline{0.831}             &\underline{0.964}                    & \underline{1.184}                  & \underline{0.969}            \\ \midrule
  DistMult &\underline{0.865}     &1.743                   &0.922                       &0.935                    &1.920                   &0.979                     &0.712                    &22.340                   &0.744                       &0.937                    &1.334                   &0.958  \\ 
  $\mathcal{L}$-DistMult &\textbf{0.955}                    &\textbf{1.134}                   &\textbf{0.988}                       &0.967                    &\textbf{1.174}                   &\textbf{0.988}                  &\textbf{0.852}                    &\textbf{2.271}                   &\textbf{0.914}                       &\textbf{0.972}                    &\textbf{1.097}                   &\textbf{0.991}    \\ \bottomrule
  \end{tabular}}
  \label{tab:result}
\end{table}

\subsection{Experimental Results and Analysis}\label{results_and_analysis}
The results on all datasets are reported in \cref{tab:result}. 
We can observe that: (1) Comparing with the original TransE and DistMult, 
our proposed $\mathcal{L}$-TransE and $\mathcal{L}$-DistMult consistently have superior performance on all datasets, 
proving that LightCAKE can greatly improve the performance of context-independent KGE models; 
(2) Comparing with all six KGE baselines, the proposed $\mathcal{L}$-TransE and $\mathcal{L}$-DistMult 
achieve substantial improvements or state-of-the-art performance on all datasets, 
showing the effectiveness of $\mathcal{L}$-TransE and $\mathcal{L}$-DistMult.

\subsection{Ablation Study}\label{ablation_study}
LightCAKE utilizes both entity context and relation context. 
How does each context affect the performance of LightCAKE? 
To answer this question, we propose model variants to conduct ablation studies on 
$\mathcal{L}$-TransE and $\mathcal{L}$-DistMult including: 
(1) the original TransE and DistMult without considering entity context and relation context; 
(2) $\mathcal{L}_{rel}$-TransE and $\mathcal{L}_{rel}$-DistMult 
that just aggregate the relation context and discard the entity  context; 
(3) $\mathcal{L}_{ent}$-TransE and $\mathcal{L}_{ent}$-DistMult that 
just aggregate the entity context and discard the relation context.

\begin{figure}[htbp]
  \centering
  \subfigure[WN18RR]{
  \label{fig:a} 
  \begin{minipage}[t]{0.30\linewidth}
  \centering
  \includegraphics[width=1\linewidth]{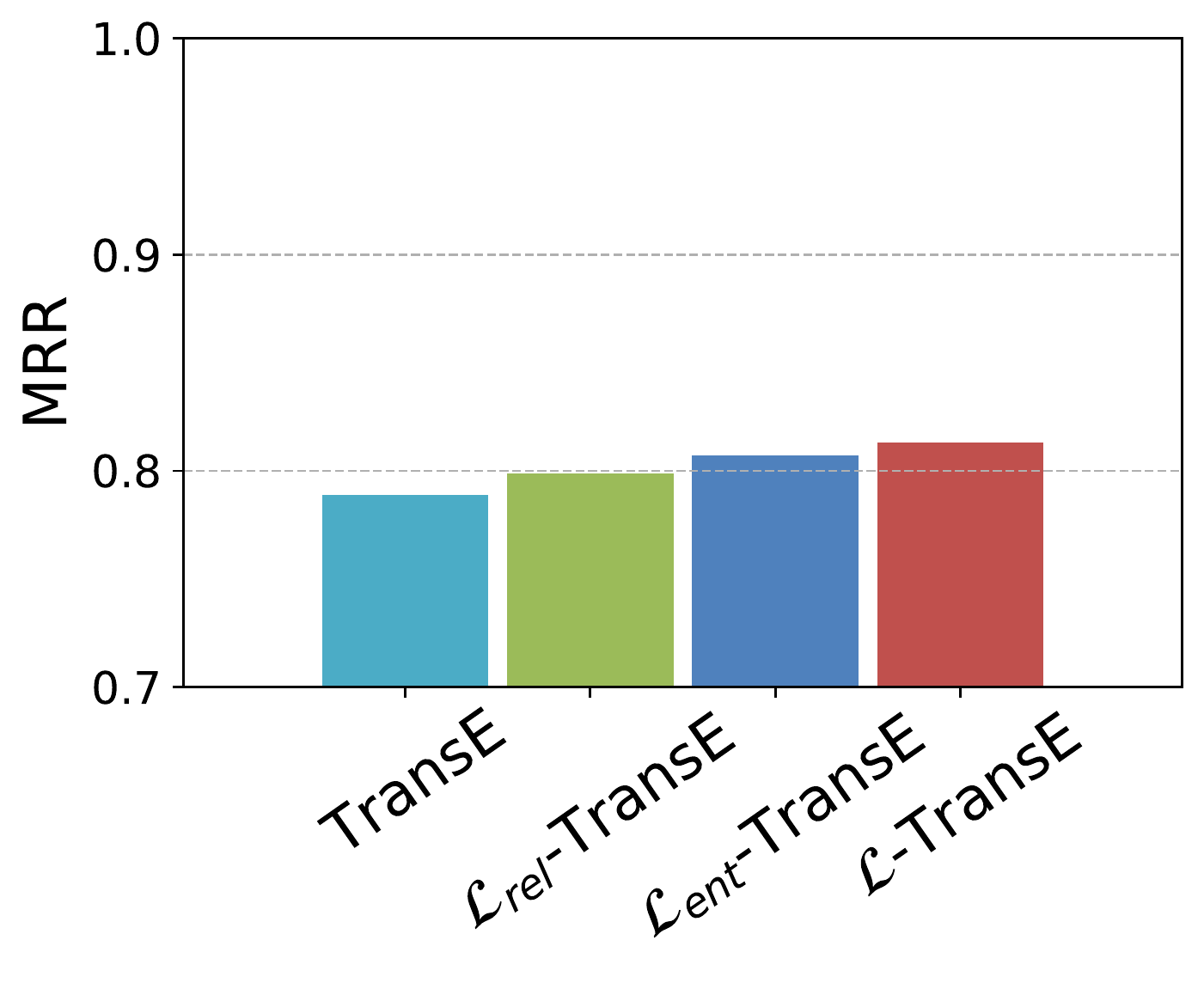}
  \end{minipage}%
  }%
  \subfigure[FB15K237]{
  \label{fig:b} 
  \begin{minipage}[t]{0.30\linewidth}
  \centering
  \includegraphics[width=1\linewidth]{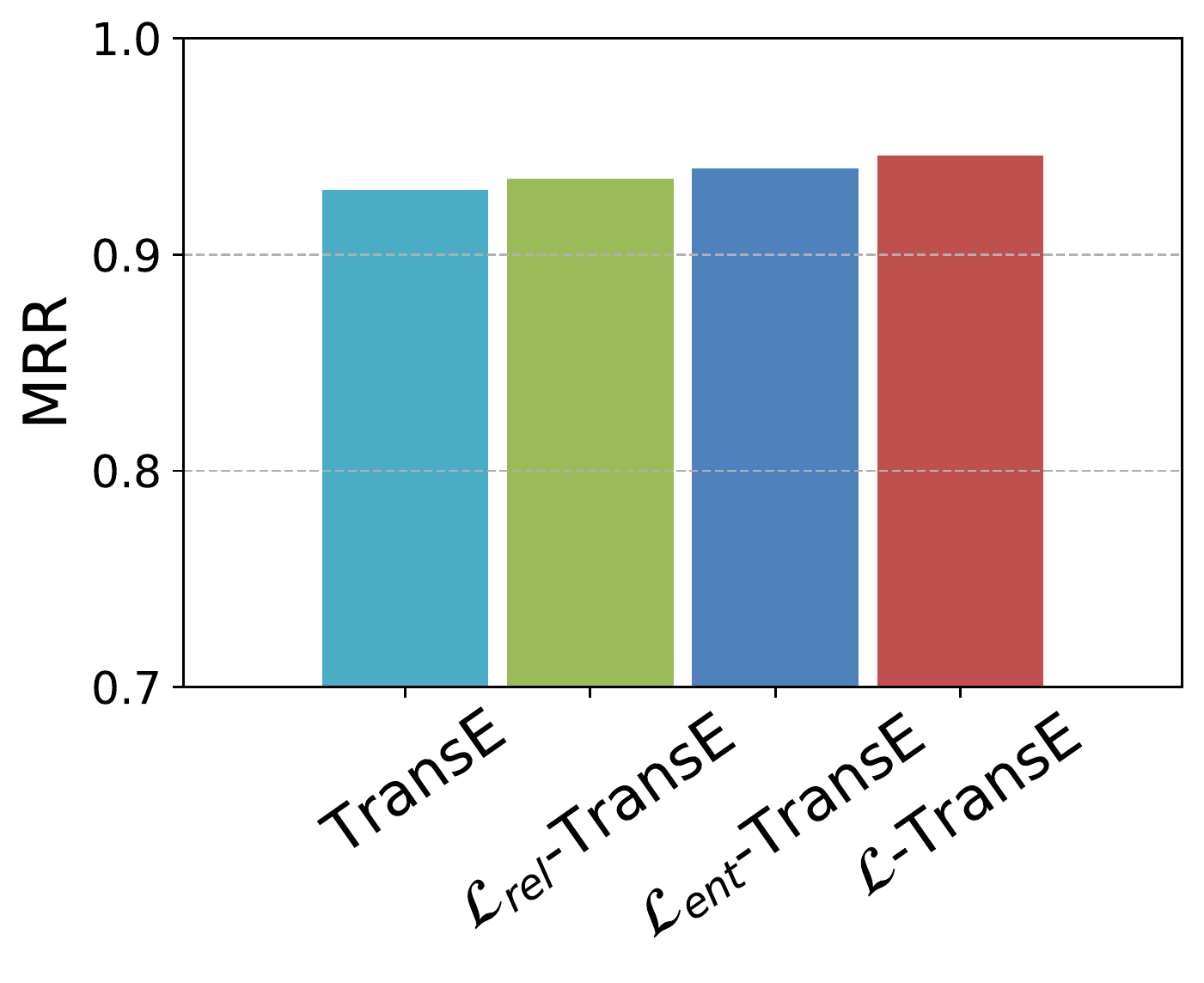}
  \end{minipage}%
  }%
  \subfigure[$\mathcal{L}$-TransE]{
  \label{fig:c} 
  \begin{minipage}[t]{0.30\linewidth}
  \centering
  \raisebox{0.23\height}{\includegraphics[width=1\linewidth]{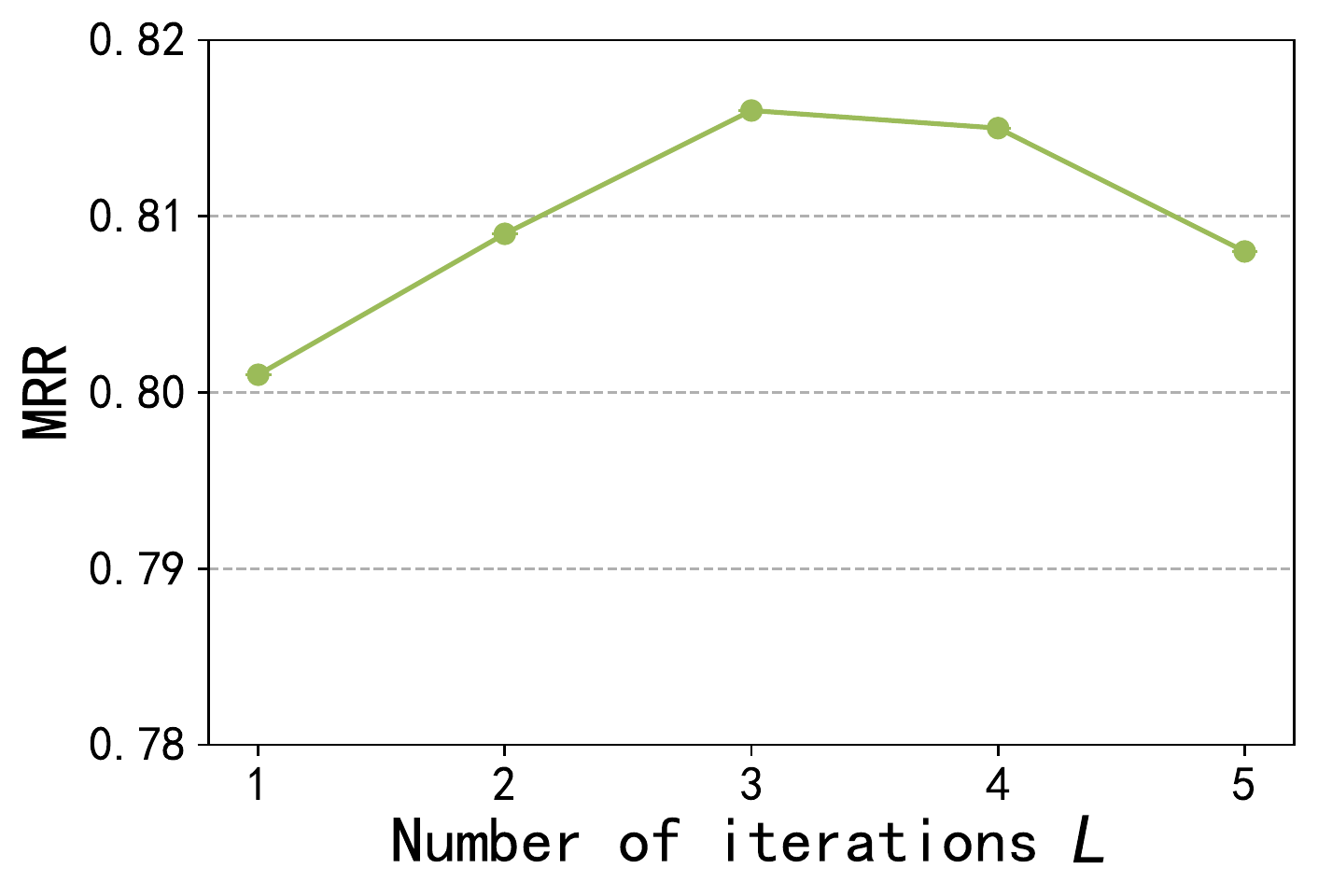}}
  \end{minipage}%
  }%
  \hfill
  \subfigure[WN18RR]{
  \label{fig:d} 
  \begin{minipage}[t]{0.30\linewidth}
  \centering
  \includegraphics[width=1\linewidth]{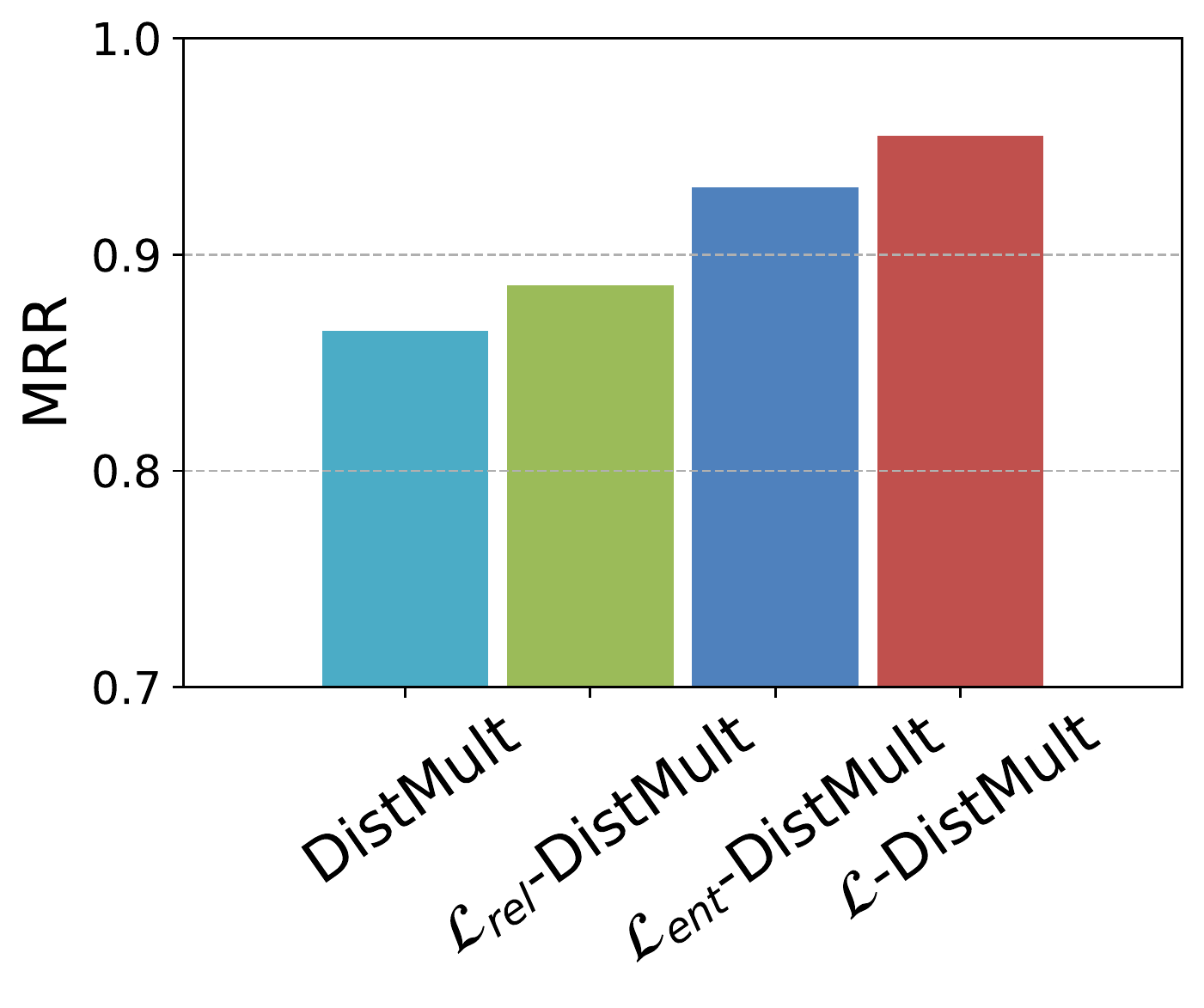}
  \end{minipage}%
  }%
  \subfigure[FB15K237]{
  \label{fig:e} 
  \begin{minipage}[t]{0.30\linewidth}
  \centering
  \includegraphics[width=1\linewidth]{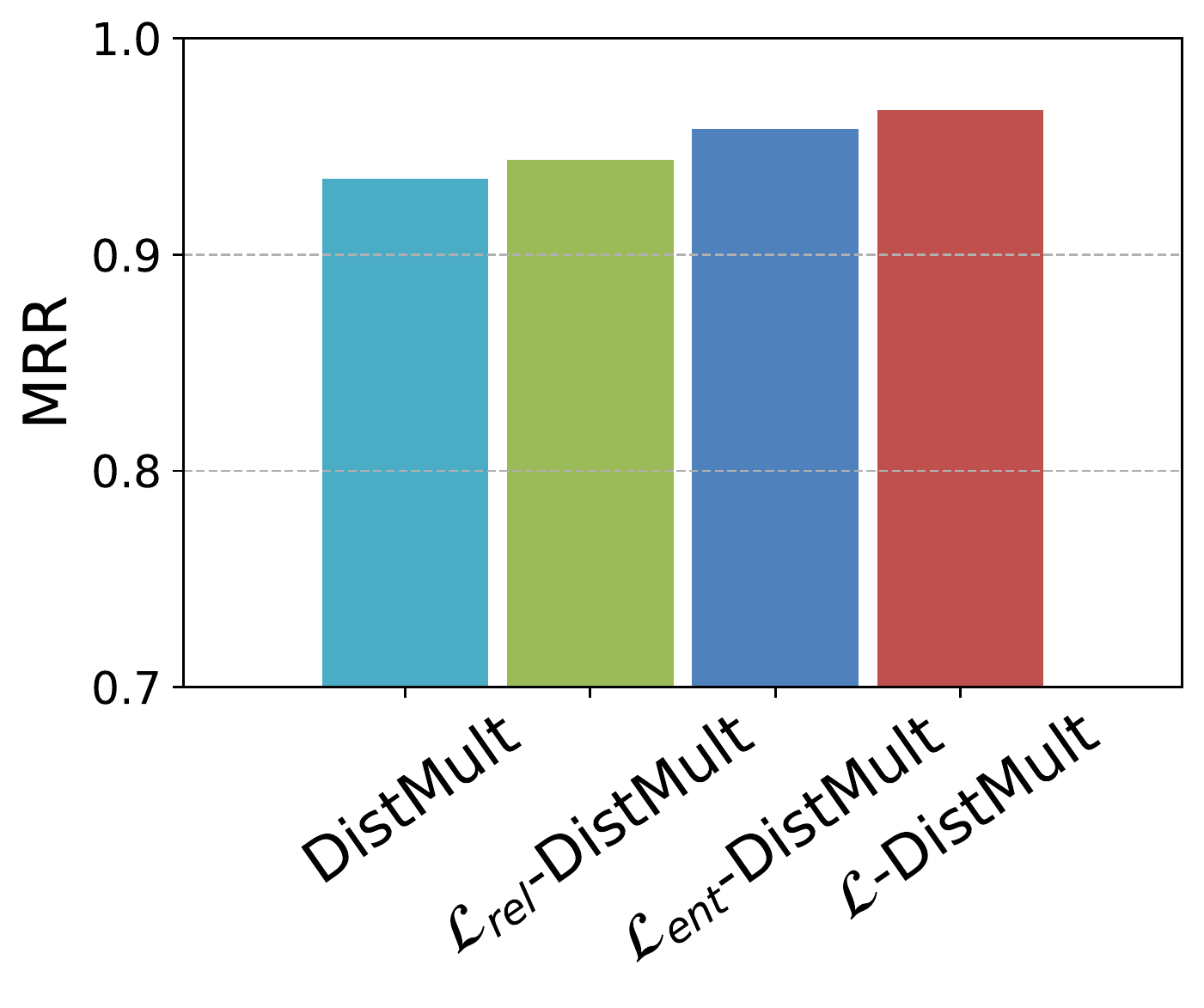}
  \end{minipage}%
  }%
  \subfigure[$\mathcal{L}$-DistMult]{
  \label{fig:f} 
  \begin{minipage}[t]{0.30\linewidth}
  \centering
  \raisebox{0.23\height}{\includegraphics[width=1\linewidth]{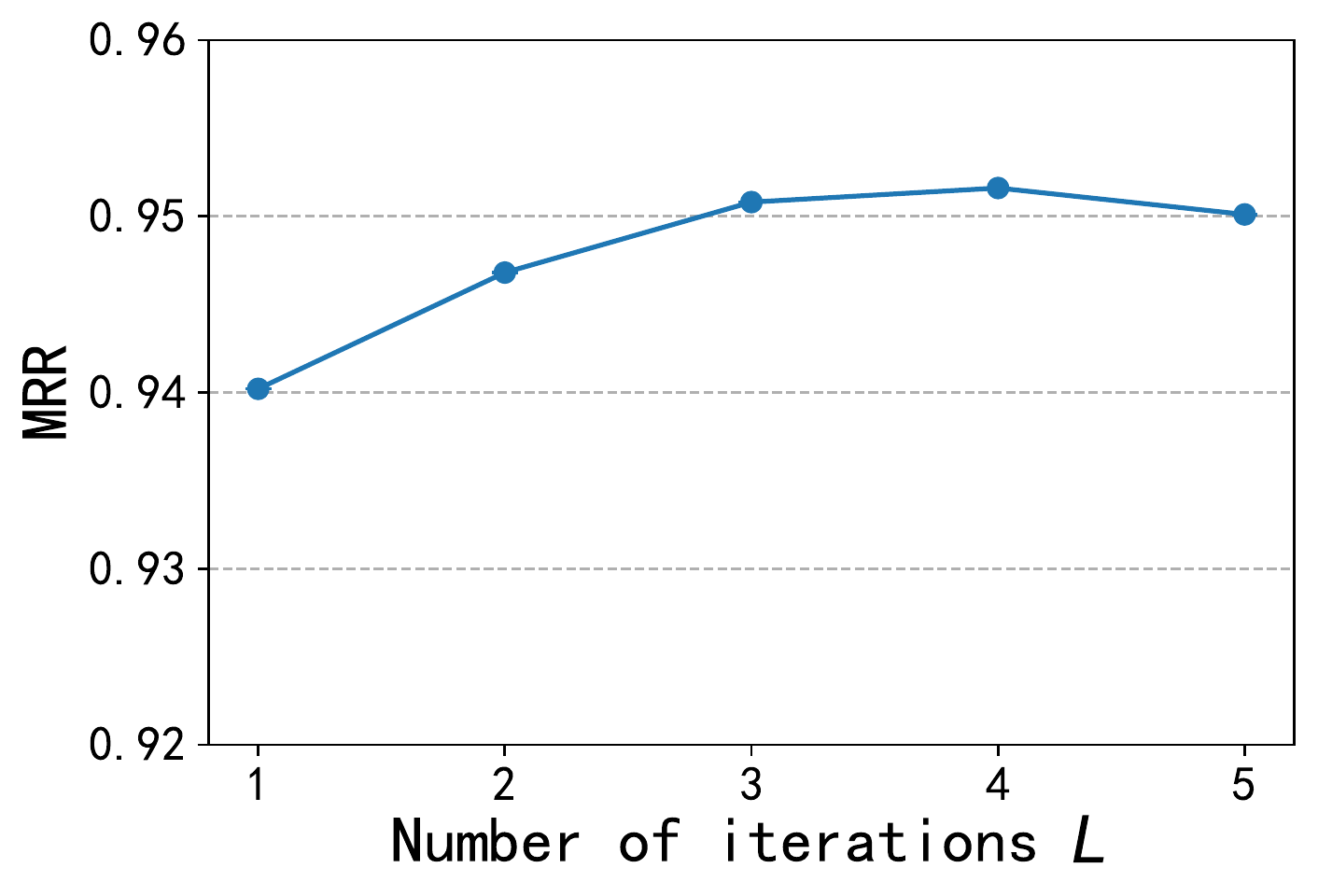}}
  \end{minipage}%
  }%
  \centering
  \caption{The performance of model variants for (a)$\mathcal{L}$-TransE and (d)$\mathcal{L}$-DistMult on  WN18RR dataset. 
  The performance of model variants for (b)$\mathcal{L}$-TransE and (e)$\mathcal{L}$-DistMult on FB15K237 dataset. 
  The performance of various $L$ for (c)$\mathcal{L}$-TransE and (f)$\mathcal{L}$-DistMult on  WN18RR dataset.}
  \label{fig:as} 
  \end{figure}

The experimental results of MRR on datasets WN18RR and FB15K237 are reported in \cref{fig:as} (a),(b),(d), and (e). 
$\mathcal{L}$-TransE and $\mathcal{L}$-DistMult 
achieve best performance compared with their corresponding model variants, 
demonstrating that integrating both entity context and relation context is most effective for KGE. 
Also, $\mathcal{L}_{rel}$-TransE and 
$\mathcal{L}_{ent}$-TransE 
are both better than TransE, 
$\mathcal{L}_{rel}$-DistMult and 
$\mathcal{L}_{ent}$-DistMult are both better than DistMult, 
indicating that entity context and relation context are both helpful for KGE. 
$\mathcal{L}_{ent}$-TransE is better than $\mathcal{L}_{rel}$-TransE 
and $\mathcal{L}_{ent}$-DistMult is better than $\mathcal{L}_{rel}$-DistMult, 
showing that entity context contributes more to improving the model performance than relation context.

\subsection{Analysis on Number of Iterations}\label{analysis_on_number_of_iterations}
In this section, 
we investigate the sensitivity of the parameter $L$, i.e., the number of iterations. 
We report the MRR on WN18RR dataset. We set that $L$ ranges from 1 to 5. 
The results of $\mathcal{L}$-TransE and $\mathcal{L}$-DistMult are shown in \cref{fig:as} (c) and (f), 
we can observe that with the growth of the number of iterations, the performance raises first and then starts to decrease slightly, 
which may due to when further contexts are involved, more uncorrelated information are integrated into embeddings. 
So properly setting the number of $L$ can help to improve the performance of our method.

\subsection{Efficiency Analysis}\label{sec:efficiency_analysis}
We evaluate the efficiency of LightCAKE by comparing it with DistMult and R-GCN. 
We investigate the difference of DistMult, R-GCN and $\mathcal{L}$-DistMult 
in the views of entity context, relation context, 
parameter quantities (space complexity), and the MRR in WN18RR dataset. 
The results are shown in \cref{tab:efficiency}.
We can observe that 
the parameter quantities of $\mathcal{L}$-DistMult are far less than R-GCN, 
that is because R-GCN use complicated matrix transformation to encode context information, 
while $\mathcal{L}$-DistMult only uses multiplication on embeddings to encode context information.
Also, both DistMult and $\mathcal{L}$-DistMult achieve better prediction results than R-GCN 
in the relation prediction task, which may because R-GCN is overfitted due to the use of too many parameters.
In summary, $\mathcal{L}$-DistMult is lighter, more efficient and more robust.

\begin{table}[htbp]
\caption{\textbf{Efficiency Analysis}.
Here, $d$ is the embedding dimension, $L$ is the number of iterations, 
$|\mathcal{E}|$ and $|\mathcal{{R}}|$ indicate the total number of entities and relations respectively.}
\centering
\begin{tabular}{ccccc}
\toprule
Models     & \begin{tabular}[c]{@{}c@{}}Entity \\ Context\end{tabular} & \begin{tabular}[c]{@{}c@{}}Relation\\ Context\end{tabular} & \begin{tabular}[c]{@{}c@{}}Space\\ Complexity\end{tabular} & MRR \\ \midrule
\small{DistMult\cite{yang2015embedding}}   & \xmark                                                         & \xmark                                                           & $\mathcal{O}(|\mathcal{E}|d + |\mathcal{{R}}|d)$                                                               &0.865     \\
\small{R-GCN\cite{schlichtkrull2018modeling}}      &\cmark                                               &    \xmark                                                         &$\mathcal{O}(L(d^{2} + |\mathcal{E}|d + |\mathcal{{R}}|d))$                                                                &0.823     \\ \midrule
\small{$\mathcal{L}$-DistMult} &\cmark                                                       &\cmark                                                         & $\mathcal{O}(L(|\mathcal{E}|d + |\mathcal{{R}}|d))$                                                               &0.955     \\ \bottomrule
\end{tabular}
\label{tab:efficiency}
\end{table}

\section{Conclusion}\label{sec:conclusion}
In this paper, we propose LightCAKE to learn context-aware knowledge graph embedding. 
LightCAKE considers both the entity context and relation context, 
and extensive experiments show its superior performance comparing with state-of-the-art KGE models. 
In addition, LightCAKE is very lightweight and efficient in aggregating context information. 
Future research will explore more possible context encoder, i.e. $\phi_{ent}$ and $\phi_{rel}$, 
and more possible scoring functions used in \cref{con:alpha}, \cref{con:beta} and \cref{con:final_p} 
to make LightCAKE more general and powerful.

\subsubsection{Acknowledgments.}
This research was supported by the Natural Science Foundation of China under Grant No. 61836013, 
the Ministry of Science and Technology Innovation Methods Special work Project under grant 2019IM020100, 
the Beijing Natural Science Foundation(4212030), 
and Beijing Nova Program of Science and Technology under Grant No. Z191100001119090. 
Zhiyuan Ning and Ziyue Qiao contribute equally to this work. 
Yi Du is the corresponding author.

%
%
%
\bibliographystyle{splncs04}
\bibliography{paper.bib}
\end{document}